\PassOptionsToPackage{table}{xcolor}
\documentclass[]{servicenow}

\usepackage{hyperref}
\usepackage[inline]{enumitem}
\usepackage{graphicx}
\usepackage{url}
\usepackage{subfig}
\usepackage{nicefrac}
\usepackage{cleveref}
\usepackage{xcolor}
\usepackage{array}
\usepackage{booktabs}
\usepackage{amsmath,amssymb,bm}
\usepackage{colortbl}
\usepackage{caption}
\usepackage{float}

\definecolor{simplexbg}{RGB}{226,239,218}
\definecolor{retrievalbg}{RGB}{218,232,252}

\title{Closing the Modality Gap via Simplex-Constrained Representations}

\author[1,2,4]{Shubham Gupta}
\author[1,2,3]{Siva Reddy}
\author[1,2,3]{Perouz Taslakian}
\author[2,4]{Valentina Zantedeschi}
\author[1,4]{Cem Subakan}

\affiliation[1]{Mila -- Qu\'ebec AI Institute}
\affiliation[2]{ServiceNow Research}
\affiliation[3]{McGill University}
\affiliation[4]{Universit\'e Laval}

\abstract{
While multimodal contrastive models like CLIP and CLAP successfully learn shared representations, their embeddings typically exhibit a persistent \emph{modality gap}. We propose addressing this separation through a purely geometric approach: changing the embedding domain from the unit sphere, $S^{d-1}$, to the probability simplex, $\Delta^{d-1}$. Although both manifolds share the same intrinsic dimension, the simplex acts as a structural prior; its nonnegative, unit-mass constraints serve as a cross-modal calibration that forces both modalities to ground themselves in the same coordinate system. By attaching lightweight softmax adapters and using Total Variation similarity, we reduce the centroid $\ell_2$ gap by 97--99\% across five multimodal retrieval benchmarks. This confirms that simplicial geometry induces genuine semantic alignment, matching or exceeding the retrieval performance of standard unconstrained models.
}

\github{\url{https://github.com/ServiceNow/retreever/tree/main/experiments/modality_gap}}
\correspondence{\email{shgup1@ulaval.ca}}

\begin{document}

\maketitle

\section{Introduction}
\label{sec:intro}

Contrastive multimodal models embed diverse data streams, such as text and audio~\cite{wu2023large,elizalde2023clap} or text and images~\cite{radford2021learning,jia2021scaling}, into a shared latent space.
Despite objectives designed to align matched pairs, these models often converge to representations in which modality-wise embedding distributions remain systematically offset, a phenomenon known as the \emph{modality gap}~\cite{liang2022mind}.
The practical consequences of this gap remain debated.
Several studies report that large modality gaps can miscalibrate cross-modal similarity, degrading retrieval and distorting zero-shot decision boundaries~\cite{zhang2023diagnosing,schrodi2024two}.
At the same time, a nonzero gap may encode modality-specific structure, and indiscriminately collapsing it can perturb within-modality geometry and harm downstream performance~\cite{liang2022mind,ramasinghe2024accept,schrodi2024two}.
Driven by this tension, a range of strategies have been proposed to manage cross-modal separation, including modifying the contrastive objective~\cite{fahim2024gapboost}, architectural fusion via shared attention layers~\cite{singh2022flava,chen2020uniter}, adversarial alignment with gradient reversal~\cite{sofer2025pull}, and alternative embeddings motivated by hierarchical structures~\cite{ramasinghe2024accept,desai2023meru}.

We take a geometric view of the modality gap and show that \emph{output geometry alone} can close it dramatically, without modifying the pretrained backbone and with fewer than 1\% trainable parameters.
Our key move is to replace the default spherical embedding domain $S^{d-1}$ with a simplex-constrained one, $\Delta^{d-1}=\{\mathbf{p}\in\mathbb{R}^d_{\ge 0}:\sum_i p_i=1\}$.
Although the sphere and simplex are both $(d\!-\!1)$-dimensional, they encode very different symmetries.
On the sphere, cosine similarity is invariant to rotations and sign flips, leaving many equivalent solutions in which modalities can match pairs while remaining distributionally separated.
On the simplex, these symmetries largely vanish: embeddings are nonnegative and sum to one, forcing both modalities to allocate a shared probability budget over the same coordinates.
With softmax adapters, the representation also becomes invariant to global logit shifts, providing an additional cross-modal normalization.

Concretely, we freeze LAION-CLAP~\cite{elizalde2023clap} for audio--text and FLAVA~\cite{singh2022flava} for image--text, attach lightweight \emph{shared} adapters, and train with an in-batch InfoNCE objective (as in DPR)\cite{oord2018representation,karpukhin2020dense}, changing only the scoring function: cosine similarity for Euclidean outputs versus negative Total Variation distance for simplex outputs produced by a softmax projection.
To isolate the effect of geometry from architectural changes, we evaluate matched-capacity baselines. 
We start from frozen backbones and train lightweight Linear and MLP adapters with Euclidean outputs, and we also train Linear/MLP adapters of the same architecture whose outputs are followed by a softmax projection to the simplex.
To disentangle output geometry from shared cross-attention, we use an Euclidean variant of ReTreever~\cite{gupta2026hierarchicalretrievalscalebridging}  (we call it \emph{XAttn}) that keeps the same learned-query cross-attention but outputs cosine-scored Euclidean embeddings, and compare it to ReTreever's simplex outputs~\cite{gupta2026hierarchicalretrievalscalebridging}.
Finally, to contrast coarse-to-fine strategies across geometries, we include MRL as an Euclidean baseline~\cite{kusupati2022matryoshka} with Retreever.

Our contributions are as follows:
(i) We propose simplex-constrained adapters for multimodal retrieval and show that they reduce the centroid $\ell_2$ modality gap by 97--99\% across five audio--text and image--text benchmarks while matching or improving NDCG@10.
(ii) We develop a geometry-adjusted gap measure, \textsc{RelGap}, which normalizes centroid separation by intra-modality spread, enabling fair comparison of modality gap across Euclidean and simplicial embedding domains.
(iii) Using matched-capacity and matched-form baselines, we isolate the effect of output geometry and show that simplex constraints (rather than additional architectural capacity) drive gap closure.
(iv) We analyze coarse-to-fine behavior and dimensional scaling, showing that Euclidean gaps grow with representation dimension whereas simplex gaps remain small across resolutions.

\section{Related Work}
\label{sec:related}

\textbf{Modality gap.}
Liang et~al.~\cite{liang2022mind} systematically characterized the modality gap in CLIP and connected it to geometric effects such as the cone effect and initialization-induced asymmetries.
Subsequent work has refined both the diagnosis and the proposed mechanisms: Fahim et~al.~\cite{fahim2024gapboost} formalize a contrastive view of the gap, Schrodi et~al.~\cite{schrodi2024two} relate it to information imbalance and compositional failures, and Yaras et~al.~\cite{yaras2025explaining} further links initialization and optimization dynamics to persistent separation.
Empirically, the gap has been associated with miscalibrated cross-modal similarities that can hurt retrieval~\cite{ma2024bridging}, distort zero-shot decision boundaries~\cite{zhang2023diagnosing}, and hinder compositional generalization~\cite{schrodi2024two}, though its impact can be regime-dependent and remains an active point of debate.

\textbf{Managing or reducing cross-modal separation.}
Prior approaches intervene either in the \emph{training objective} or the \emph{architecture}.
On the objective side, methods modify contrastive learning to explicitly encourage alignment or reweight supervision~\cite{fahim2024gapboost}, use adversarial signals such as gradient reversal to suppress modality-specific features~\cite{sofer2025pull}, or propose structured/group-wise alignment strategies~\cite{grassucci2026closing}.
On the architectural side, fusion models introduce shared cross-modal processing, e.g., shared transformer layers or attention blocks, as in UNITER~\cite{chen2020uniter}, FLAVA~\cite{singh2022flava}, VATT~\cite{akbari2021vatt}, and BLIP-2~\cite{li2023blip}.
A complementary line adds lightweight adapters or projection heads on top of frozen backbones (e.g., CLIP-Adapter~\cite{gao2024clip} and parameter-efficient tuning methods such as LoRA~\cite{houlsby2019parameter,hu2022lora}), aiming to improve alignment with minimal trainable parameters.
Finally, hyperbolic spaces have been explored to better capture hierarchical structure~\cite{ramasinghe2024accept,desai2023meru}.

\textbf{Simplex and distributional representations.}
Simplex-valued representations naturally model nonnegative mixture weights, with roots in compositional data analysis~\cite{aitchison1982statistical} and topic models such as LDA~\cite{blei2003latent}.
In deep learning, simplex/Dirichlet parameterizations appear in uncertainty modeling~\cite{sensoy2018evidential} and in neural collapse, where late-stage features exhibit simplex-like structure~\cite{papyan2020prevalence}.
Most relevant here, ReTreever uses softmax-normalized coarse-to-fine simplex embeddings for retrieval and is competitive with Euclidean multi-resolution baselines such as MRL~\cite{gupta2026hierarchicalretrievalscalebridging,kusupati2022matryoshka}.

\section{Method}
\label{sec:method}

\subsection{Adapters and output domains}
\label{sec:simplex}

Let $f(\cdot)$ denote a frozen multimodal encoder (audio--text or image--text). For an input $u$ (text or audio/image), the encoder produces a pooled embedding $\mathbf{h}_u \in \mathbb{R}^d$ 
.
We attach a lightweight \emph{shared} adapter $g_\theta$ (either linear or a 2-layer MLP) to obtain logits $\mathbf{z}_u = g_\theta(\mathbf{h}_u)$.
We study two output domains using the same adapter architecture, isolating the effect of geometry from adapter capacity. \textbf{Euclidean (spherical) outputs.}
We form normalized embeddings $\mathbf{e}_u=\mathbf{z}_u/\|\mathbf{z}_u\|_2 \in S^{d-1}$ and score pairs with cosine similarity,
$s_{\mathrm{cos}}(u,v)=\mathbf{e}_u^\top \mathbf{e}_v$. \textbf{Simplex-constrained outputs.}
We project logits to the simplex via softmax, $\mathbf{p}_u=\mathrm{softmax}(\mathbf{z}_u)\in\Delta^{d-1}$, and score pairs with a Total Variation--based score,
$
s_{\mathrm{TV}}(u,v) \;=\; -\mathrm{TV}(\mathbf{p}_u,\mathbf{p}_v)
\;=\; -\tfrac{1}{2}\|\mathbf{p}_u-\mathbf{p}_v\|_1
$. Beyond Linear/MLP adapters, we evaluate cross-attention and multi-resolution baselines. 
\textbf{XAttn} learns 100 shared 1024-d queries that cross-attend to token-level representations and are mean-pooled into a cosine-scored Euclidean embedding, while \textbf{ReTreever} uses similar mechanism to produce multi-resolution simplex embeddings via softmax~\cite{gupta2026hierarchicalretrievalscalebridging}. 
We also train \textbf{MRL}~\cite{kusupati2022matryoshka}, which jointly optimizes multiple prefix slices of a single Euclidean embedding under the same retrieval objective.
\subsection{Training objective}
\label{sec:loss}

We train the adapter(s) using a symmetric in-batch contrastive loss (DPR-style)~\cite{karpukhin2020dense}.
Let a minibatch contain paired examples $\{(t_i, x_i)\}_{i=1}^B$, where $t$ is text and $x$ is audio or image.
Given a scoring function $s(\cdot,\cdot)$ (cosine for Euclidean outputs, negative TV for simplex), we optimize $\mathcal{L}=\tfrac{1}{2}\big(\mathcal{L}^{t\to x}_{\mathrm{NCE}}+\mathcal{L}^{x\to t}_{\mathrm{NCE}}\big)$, where: 
$
\mathcal{L}^{t\to x}_{\mathrm{NCE}}
= -\frac{1}{B}\sum_{i=1}^{B}
\log
\frac{\exp(s(t_i,x_i))}
{\sum_{j=1}^{B}\exp(s(t_i,x_j))}$, and $
\mathcal{L}^{x\to t}_{\mathrm{NCE}}
= -\frac{1}{B}\sum_{i=1}^{B}
\log
\frac{\exp(s(x_i,t_i))}
{\sum_{j=1}^{B}\exp(s(x_i,t_j))}
$

\subsection{Sphere vs.\ simplex: symmetries and distance scaling}
\label{sec:geometry}

Switching the output domain from the sphere to the simplex changes both (i) the invariances of the representation and (ii) the baseline behavior of distances in high dimensions. These geometric properties are important for interpreting modality gaps across domains.

\textbf{Symmetries and calibration.}
With Euclidean outputs, we normalize embeddings and use cosine similarity, so representations lie on the unit sphere $S^{d-1}$.
Cosine similarity is invariant to orthogonal transformations (rotations/reflections) and sign flips, which implies a large family of equivalent solutions: matched pairs can be aligned while modality-wise distributions remain offset.
With simplex outputs, embeddings are constrained to $\Delta^{d-1}=\{\mathbf{p}\in\mathbb{R}^d_{\ge 0}:\sum_i p_i=1\}$, which removes most of this freedom by enforcing nonnegativity and unit mass.
As a result, both modalities must express representations as comparable allocations of a shared probability budget over the same coordinates.
When the simplex embedding is produced via $\mathbf{p}=\mathrm{softmax}(\mathbf{z})$, the representation is also invariant to global logit shifts: $\mathrm{softmax}(\mathbf{z}+c\mathbf{1})=\mathrm{softmax}(\mathbf{z})$, providing an additional cross-modal normalization.

\textbf{Baseline distances.}
On $S^{d-1}$, random unit vectors become nearly orthogonal as $d$ increases, and pairwise distances concentrate around a constant: for $\mathbf{x},\mathbf{y}\sim\mathrm{Unif}(S^{d-1})$, $\mathbb{E}\|\mathbf{x}-\mathbf{y}\|_2^2=2$ and $\|\mathbf{x}-\mathbf{y}\|_2\approx\sqrt{2}$ in high dimension.
On the simplex, typical distances depend on how \emph{dense} the probability vectors are.
A convenient reference model is the symmetric Dirichlet distribution $\mathbf{p},\mathbf{q}\stackrel{\mathrm{iid}}{\sim}\mathrm{Dir}(\alpha,\ldots,\alpha)$, for which
$\mathbb{E}\|\mathbf{p}-\mathbf{q}\|_2^2
= \frac{2(\alpha+1)}{d\alpha+1} - \frac{2}{d}$. This highlights two regimes: (i) \emph{dense} simplex embeddings (fixed $\alpha$) where typical distances shrink as $\mathcal{O}(1/\sqrt{d})$, and (ii) \emph{sparse} embeddings ($\alpha=c/d$ with $d\alpha=c$ held constant) where distances plateau at $\approx 2/(c+1)$, analogous to the spherical concentration phenomenon.

\textbf{Estimating the regime and comparing gaps across domains.}
To connect this characterization to learned simplex embeddings, we estimate a concentration parameter via moment matching.
Let $s=\mathbb{E}\|\mathbf{p}\|_2^2$ for learned $\mathbf{p}\in\Delta^{d-1}$; then
$
\hat{\alpha}=\frac{1-s}{sd-1}.
$
We use $\hat{\alpha}$ to determine whether learned simplex representations lie in a dense (contracting) or sparse (plateauing) regime as resolution/dimension increases (see \S\ref{sec:coarse}).
Finally, because raw centroid distances can be influenced by the baseline distance scale of the embedding domain, we report a geometry-adjusted \textsc{RelGap} metric (Eq.~\ref{eq:relgap}) that normalizes centroid separation by intra-modality spread, enabling fairer gap comparisons between Euclidean and simplex outputs.

\begin{figure}[t]
\centering
\includegraphics[width=\columnwidth]{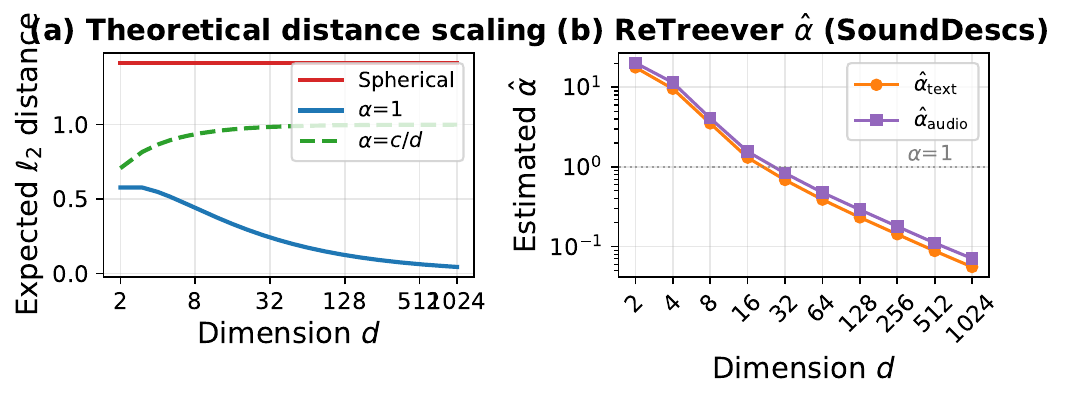}
\caption{\textbf{(a)} Expected $\ell_2$ distance vs.\ dimension. Sphere: plateau at $\sqrt{2}$. Dense simplex ($\alpha=1$): shrinks as $\mathcal{O}(1/\sqrt{d})$. Sparse simplex ($\alpha=c/d$): plateau at $\sqrt{2/(c+1)}$.
\textbf{(b)} Dirichlet $\hat{\alpha}$ estimated from learned ReTreever embeddings (SoundDescs). As dimension grows, $\hat{\alpha}\ll 1$, placing embeddings in the sparse regime where distances plateau rather than contract.}
\label{fig:theory}
\end{figure}

\begin{table*}[t]
\caption{Audio-text modality gap and retrieval. Sfx = softmax. Simplex methods \colorbox{simplexbg}{shaded}. Modality gap measured as Centroid $\ell_2$, RelGap (RG, Eq.~\ref{eq:relgap}),  silhouette (Sil) (lower is better ($\downarrow$)),  Retrieval as NDCG@10 (higher is better ($\uparrow$)). Best per column in \textbf{bold}. $^\dagger$Lin.+Sfx and MLP+Sfx: per dataset, the better of the paper's configuration (L2-normalised adapter input, no logit norm) and the same adapter trained on un-normalised input with a LayerNorm on the logits before the softmax; the other rows are unchanged.}
\label{tab:audio}
\centering
\footnotesize
\setlength{\tabcolsep}{2pt}
\begin{tabular}{@{}l !{\vrule width 0.6pt} ccc !{\vrule width 0.3pt} cc !{\vrule width 0.6pt} ccc !{\vrule width 0.3pt} cc !{\vrule width 0.6pt} ccc !{\vrule width 0.3pt} cc @{}}
\toprule
& \multicolumn{5}{c!{\vrule width 0.6pt}}{\textbf{Clotho}} & \multicolumn{5}{c!{\vrule width 0.6pt}}{\textbf{SoundDescs}} & \multicolumn{5}{c}{\textbf{AudioCaps}} \\
\cmidrule(lr){2-6} \cmidrule(lr){7-11} \cmidrule(lr){12-16}
& \multicolumn{3}{c!{\vrule width 0.3pt}}{Gap} & \multicolumn{2}{c!{\vrule width 0.6pt}}{\cellcolor{retrievalbg}Retrieval}
& \multicolumn{3}{c!{\vrule width 0.3pt}}{Gap} & \multicolumn{2}{c!{\vrule width 0.6pt}}{\cellcolor{retrievalbg}Retrieval}
& \multicolumn{3}{c!{\vrule width 0.3pt}}{Gap} & \multicolumn{2}{c}{\cellcolor{retrievalbg}Retrieval} \\
\cmidrule(lr){2-4} \cmidrule(lr){5-6} \cmidrule(lr){7-9} \cmidrule(lr){10-11} \cmidrule(lr){12-14} \cmidrule(lr){15-16}
\textbf{Adapter} & {$\ell_2${\tiny$\downarrow$}} & {RG{\tiny$\downarrow$}} & {Sil{\tiny$\downarrow$}} & \cellcolor{retrievalbg}{T2A{\tiny$\uparrow$}} & \cellcolor{retrievalbg}{A2T{\tiny$\uparrow$}} & {$\ell_2${\tiny$\downarrow$}} & {RG{\tiny$\downarrow$}} & {Sil{\tiny$\downarrow$}} & \cellcolor{retrievalbg}{T2A{\tiny$\uparrow$}} & \cellcolor{retrievalbg}{A2T{\tiny$\uparrow$}} & {$\ell_2${\tiny$\downarrow$}} & {RG{\tiny$\downarrow$}} & {Sil{\tiny$\downarrow$}} & \cellcolor{retrievalbg}{T2A{\tiny$\uparrow$}} & \cellcolor{retrievalbg}{A2T{\tiny$\uparrow$}} \\
\midrule
CLAP        & .640 & .748 & .197 & \cellcolor{retrievalbg}.259 & \cellcolor{retrievalbg}.436 & .488 & .521 & .117 & \cellcolor{retrievalbg}.273 & \cellcolor{retrievalbg}.250 & .363 & .380 & .067 & \cellcolor{retrievalbg}.596 & \cellcolor{retrievalbg}\textbf{.507} \\
Linear      & .514 & .689 & .167 & \cellcolor{retrievalbg}.373 & \cellcolor{retrievalbg}.453 & .458 & .554 & .133 & \cellcolor{retrievalbg}.410 & \cellcolor{retrievalbg}.418 & .366 & .468 & .139 & \cellcolor{retrievalbg}.584 & \cellcolor{retrievalbg}.439 \\
MLP         & .323 & .353 & .064 & \cellcolor{retrievalbg}.358 & \cellcolor{retrievalbg}.436 & .220 & .264 & .033 & \cellcolor{retrievalbg}.520 & \cellcolor{retrievalbg}.514 & .266 & .285 & .049 & \cellcolor{retrievalbg}.587 & \cellcolor{retrievalbg}.448 \\
XAttn       & .450 & .443 & .126 & \cellcolor{retrievalbg}.353 & \cellcolor{retrievalbg}.432 & .309 & .331 & .051 & \cellcolor{retrievalbg}.464 & \cellcolor{retrievalbg}.456 & .297 & .321 & .053 & \cellcolor{retrievalbg}.592 & \cellcolor{retrievalbg}.452 \\
\rowcolor{simplexbg}
Lin.+Sfx$^\dagger$ & \textbf{.007} & .468 & .127 & \textbf{.382} & \textbf{.470} & .021 & .196 & .022 & .437 & .436 & .012 & .217 & .030 & .584 & .440 \\
\rowcolor{simplexbg}
MLP+Sfx$^\dagger$ & .011 & .336 & .089 & .361 & .438 & \textbf{.010} & .179 & .018 & .507 & .499 & \textbf{.006} & .180 & .012 & \textbf{.597} & .451 \\
\rowcolor{simplexbg}
ReTreever   & .020 & \textbf{.181} & \textbf{.026} & .381 & .464 & .012 & \textbf{.097} & \textbf{.006} & \textbf{.535} & \textbf{.542} & .011 & \textbf{.113} & \textbf{.000} & .590 & .444 \\
\bottomrule
\end{tabular}
\end{table*}

\begin{figure*}[t]
\begin{minipage}[b]{0.62\textwidth}
\centering
\footnotesize
\setlength{\tabcolsep}{3.5pt}
\begin{tabular}{@{}l !{\vrule width 0.6pt} ccc !{\vrule width 0.3pt} cc !{\vrule width 0.6pt} ccc !{\vrule width 0.3pt} cc @{}}
\toprule
& \multicolumn{5}{c!{\vrule width 0.6pt}}{\textbf{COCO}} & \multicolumn{5}{c}{\textbf{Flickr30k}} \\
\cmidrule(lr){2-6} \cmidrule(lr){7-11}
& \multicolumn{3}{c!{\vrule width 0.3pt}}{Gap} & \multicolumn{2}{c!{\vrule width 0.6pt}}{\cellcolor{retrievalbg}Retrieval}
& \multicolumn{3}{c!{\vrule width 0.3pt}}{Gap} & \multicolumn{2}{c}{\cellcolor{retrievalbg}Retrieval} \\
\cmidrule(lr){2-4} \cmidrule(lr){5-6} \cmidrule(lr){7-9} \cmidrule(lr){10-11}
\textbf{Adapter} & {$\ell_2${\tiny$\downarrow$}} & {RG{\tiny$\downarrow$}} & {Sil.{\tiny$\downarrow$}} & \cellcolor{retrievalbg}{T2I{\tiny$\uparrow$}} & \cellcolor{retrievalbg}{I2T{\tiny$\uparrow$}} & {$\ell_2${\tiny$\downarrow$}} & {RG{\tiny$\downarrow$}} & {Sil.{\tiny$\downarrow$}} & \cellcolor{retrievalbg}{T2I{\tiny$\uparrow$}} & \cellcolor{retrievalbg}{I2T{\tiny$\uparrow$}} \\
\midrule
FLAVA       & .810 & 1.00 & .336 & \cellcolor{retrievalbg}.630 & \cellcolor{retrievalbg}.529 & .731 & .891 & .285 & \cellcolor{retrievalbg}.801 & \cellcolor{retrievalbg}.695 \\
Linear      & .647 & .732 & .212 & \cellcolor{retrievalbg}\textbf{.711} & \cellcolor{retrievalbg}\textbf{.644} & .543 & .598 & .151 & \cellcolor{retrievalbg}\textbf{.858} & \cellcolor{retrievalbg}.817 \\
MLP         & .411 & .438 & .088 & \cellcolor{retrievalbg}.688 & \cellcolor{retrievalbg}.615 & .246 & .277 & .036 & \cellcolor{retrievalbg}.828 & \cellcolor{retrievalbg}.792 \\
XAttn       & .615 & .666 & .182 & \cellcolor{retrievalbg}.708 & \cellcolor{retrievalbg}.641 & .527 & .563 & .136 & \cellcolor{retrievalbg}.857 & \cellcolor{retrievalbg}\textbf{.819} \\
\rowcolor{simplexbg}
Lin.+Sfx$^\dagger$ & .021 & .475 & .113 & .691 & .593 & .017 & .328 & .071 & .844 & .787 \\
\rowcolor{simplexbg}
MLP+Sfx$^\dagger$ & .012 & .239 & .029 & .686 & .606 & .022 & .324 & .052 & .824 & .783 \\
\rowcolor{simplexbg}
ReTreever   & \textbf{.007} & \textbf{.221} & \textbf{.026} & .694 & .613 & \textbf{.005} & \textbf{.129} & \textbf{.008} & .837 & .799 \\
\bottomrule
\end{tabular}
\captionof{table}{Image-text modality gap and retrieval (FLAVA encoder). Sfx = softmax. Simplex methods \colorbox{simplexbg}{shaded}. Modality Gap measured by Centroid $\ell_2$, RelGap (RG, Eq.~\ref{eq:relgap}),  silhouette (Sil) and Retrieval as NDCG@10. Best per column in \textbf{bold}. $^\dagger$Lin.+Sfx and MLP+Sfx: the same adapters trained on un-normalised adapter input (no L2 normalisation), no norm on the logits; the other rows are unchanged.}
\label{tab:image}
\end{minipage}\hfill
\begin{minipage}[b]{0.36\textwidth}
\centering
\includegraphics[width=\textwidth]{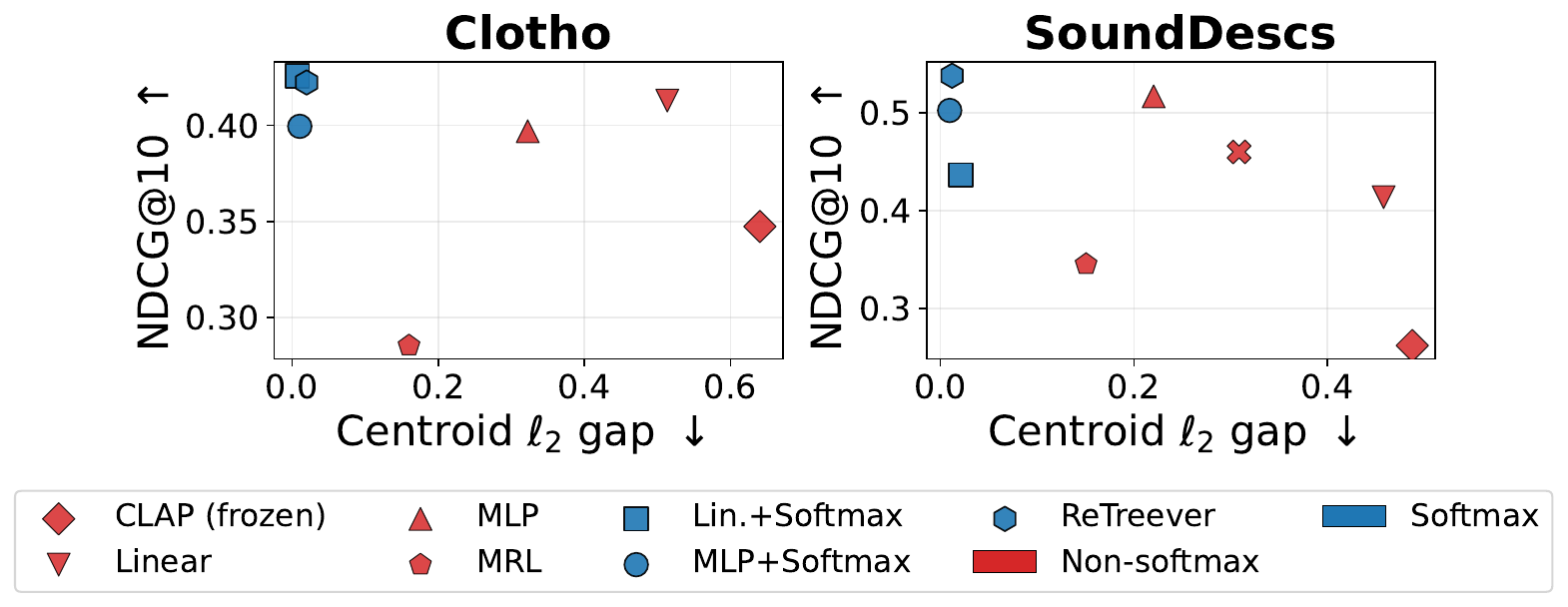}
\captionof{figure}{Modality gap (centroid $\ell_2$) vs.\ avg.\ retrieval NDCG@10. Softmax methods (\textcolor{blue}{blue}) cluster at low gap with competitive retrieval; non-softmax (\textcolor{red}{red}) spread across higher gaps.}
\label{fig:scatter}
\end{minipage}
\end{figure*}

\begin{figure}[t]
\centering
\includegraphics[width=\columnwidth]{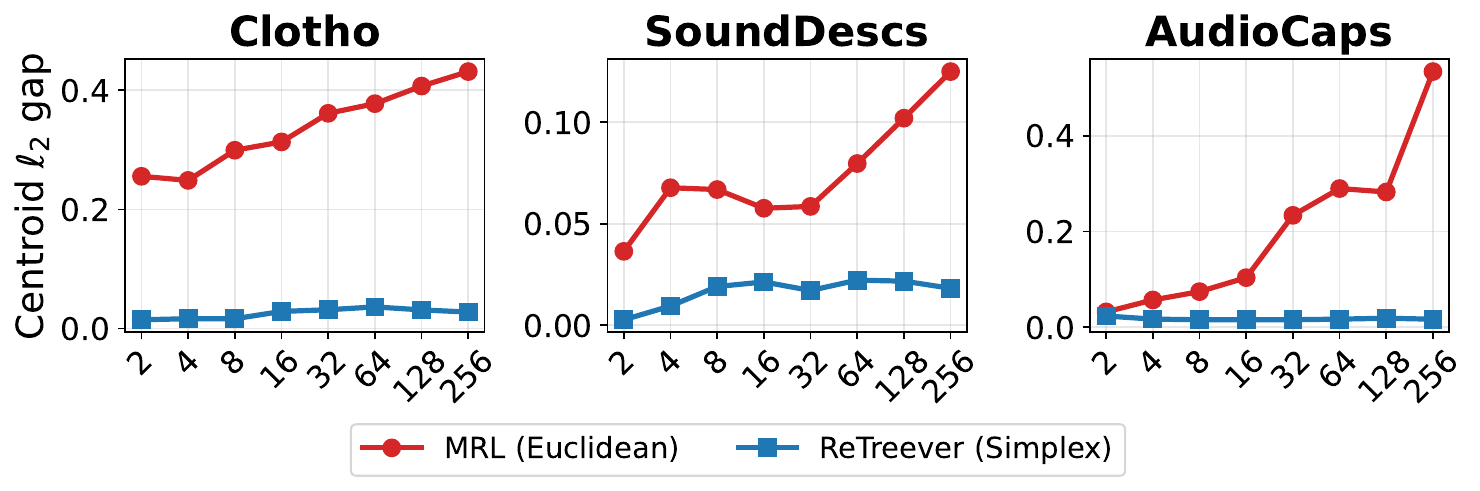}
\caption{Coarse-to-fine modality gap: MRL (Euclidean, \textcolor{red}{red}) vs.\ ReTreever (simplex, \textcolor{blue}{blue}) across representation dimensions. Euclidean gaps grow; simplex gaps stay flat.}
\label{fig:coarse}
\end{figure}

\begin{figure}[t]
\centering
\includegraphics[width=\columnwidth]{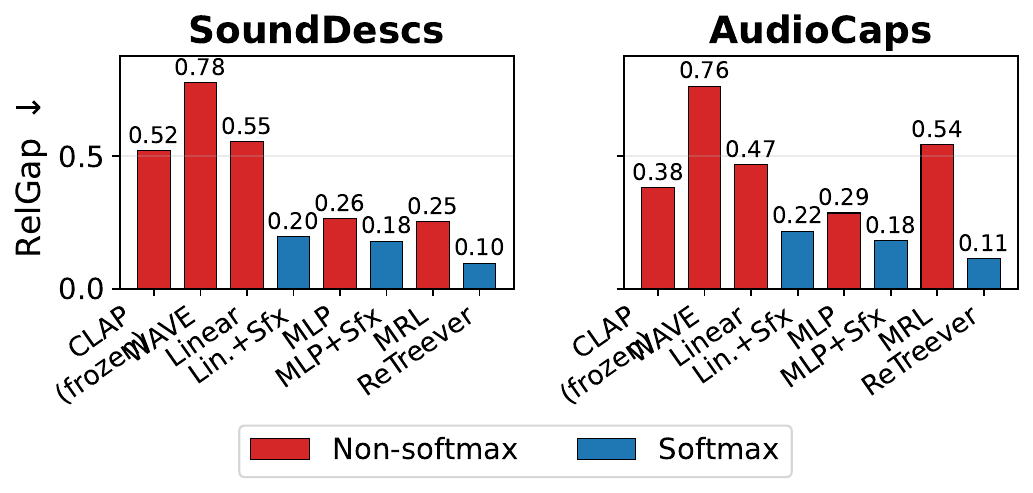}
\caption{Relative gap (RelGap, Eq.~\ref{eq:relgap}) across adapter types on SoundDescs (left) and AudioCaps (right). Non-softmax methods (\textcolor{red}{red}) vs.\ softmax methods (\textcolor{blue}{blue}). Lower is better. Softmax methods consistently reduce RelGap.}
\label{fig:relgap_bars}
\end{figure}

\subsection{Baselines}
\label{sec:baselines}

\textbf{Linear/MLP Adapters}
We use Linear and 2-layer MLP adapters in Euclidean space, and the \emph{same} adapters followed by a softmax projection to the simplex.
These paired baselines keep architecture fixed and isolate the effect of output geometry.

\textbf{Cross-attention adapters}
XAttn uses ReTreever's shared cross-attention mechanism, in which a fixed set of learned queries attends over token representations, but produces Euclidean embeddings scored with cosine similarity.
ReTreever uses the same cross-attention backbone but outputs simplex-valued representations via softmax and is scored with a TV-based measure~\cite{gupta2026hierarchicalretrievalscalebridging}.
Comparing XAttn to ReTreever isolates the effect of simplex output constraints from the effect of shared cross-attention.

\textbf{Coarse-to-fine representations}
MRL~\cite{kusupati2022matryoshka} provides a multi-resolution Euclidean baseline (Matryoshka embeddings) enabling coarse-to-fine retrieval in Euclidean space.
ReTreever provides the corresponding coarse-to-fine retrieval mechanism on the simplex~\cite{gupta2026hierarchicalretrievalscalebridging}.
We use this pairing to study how geometry affects gap scaling with effective dimension (see \S\ref{sec:coarse}).

\section{Experimental Setup}
\label{sec:experiments}

\textbf{Datasets.}
We evaluate on five cross-modal retrieval benchmarks spanning audio--text and image--text alignment.
For audio--text, we use Clotho~\cite{drossos2020clotho}, AudioCaps~\cite{kim2019audiocaps}, and SoundDescs~\cite{koepke2022audio}.
For image--text, we use MS-COCO Captions~\cite{lin2014microsoft} and Flickr30k~\cite{young2014image}.
All datasets consist of paired items with bidirectional retrieval (text$\leftrightarrow$audio or text$\leftrightarrow$image).

\textbf{Backbones and adapters.}
For audio--text experiments we freeze LAION-CLAP~\cite{laionclap2023}, and for image--text we freeze FLAVA~\cite{singh2022flava}.
On top of each frozen backbone we evaluate seven configurations: the frozen encoder without any adapter; Linear or MLP adapters with Euclidean outputs scored by cosine similarity; Linear/MLP adapters followed by a softmax projection to the simplex and scored with a TV-based measure; XAttn (shared cross-attention with Euclidean outputs and cosine similarity); and ReTreever (shared cross-attention with multi-resolution simplex outputs)~\cite{gupta2026hierarchicalretrievalscalebridging}.
For coarse-to-fine analysis (\S\ref{sec:coarse}), we additionally compare MRL~\cite{kusupati2022matryoshka} (multi-resolution Euclidean) against ReTreever. All adapters are trained for 200K steps with AdamW (lr $=4\times10^{-4}$), batch size 64, and a learnable temperature $\tau$, totaling fewer than 1\% trainable parameters.

\textbf{Metrics.}
We measure the modality gap using centroid $\ell_2$, silhouette~\cite{rousseeuw1987silhouettes}, and a geometry-adjusted relative gap metric.
Centroid $\ell_2$ is $\|\boldsymbol{\mu}_A-\boldsymbol{\mu}_B\|_2$, where $\boldsymbol{\mu}_A$ and $\boldsymbol{\mu}_B$ are modality centroids.
Silhouette summarizes cluster separability via the standard per-point coefficient $(b-a)/\max(a,b)$, averaged over points, where $a$ is the mean intra-modality distance and $b$ is the mean nearest-modality distance.
To make gaps comparable across Euclidean and simplex outputs, we report \textsc{RelGap}, which normalizes centroid separation by intra-modality spread:
\begin{equation}
\text{RelGap}(A,B)=
\frac{\|\boldsymbol{\mu}_A-\boldsymbol{\mu}_B\|_2}
{\sqrt{\tfrac{1}{2}\!\left(\mathbb{E}\|x_A-\boldsymbol{\mu}_A\|_2^2+\mathbb{E}\|x_B-\boldsymbol{\mu}_B\|_2^2\right)} }.
\label{eq:relgap}
\end{equation}
We report retrieval quality using NDCG@10 for both directions (text$\to$audio/audio$\to$text or text$\to$image/image$\to$text).

\section{Results}
\label{sec:results}

\subsection{Simplex projection collapses the gap while preserving retrieval}
Tables~\ref{tab:audio}--\ref{tab:image} and Figure~\ref{fig:scatter} summarize results across five benchmarks (audio--text and image--text).
Across all datasets, switching an otherwise identical adapter from Euclidean output to a softmax projection onto the simplex collapses the modality gap by two orders of magnitude, without sacrificing retrieval.

On frozen backbones, centroid $\ell_2$ gaps are large (0.36--0.81 across datasets).
Adding a softmax to the same Linear/MLP adapter reduces centroid $\ell_2$ to 0.005--0.020, corresponding to a 97--99\% reduction.
For example, COCO drops from 0.810 (frozen) to 0.009 (MLP+Sfx), and Flickr30k drops from 0.731 (frozen) to 0.007 (MLP+Sfx).
Figure~\ref{fig:scatter} shows the same pattern geometrically: softmax (simplex) methods cluster at low gap while remaining competitive in average NDCG@10.

Because the absolute distance scale differs across output domains, we also report \textsc{RelGap} (Eq.~\ref{eq:relgap}), which normalizes centroid separation by intra-modality spread.
Simplex methods consistently achieve the lowest \textsc{RelGap}, indicating that cross-modal alignment improves beyond any uniform rescaling of distances.
For instance, on SoundDescs, \textsc{RelGap} drops from 0.521 (frozen) to 0.179 (MLP+Sfx) and to 0.097 (ReTreever), and on AudioCaps it drops from 0.380 (frozen) to 0.180 (MLP+Sfx) and to 0.113 (ReTreever).

Finally, gap reduction does not trade off with retrieval quality.
On Clotho, Lin.+Sfx improves T2A NDCG@10 from 0.259 to 0.382, and on SoundDescs, ReTreever achieves the best retrieval scores (0.535/0.542) while maintaining a very small gap.
Overall, simplex constraints remove modality-specific distributional offsets while preserving (and sometimes improving) the neighborhood structure relevant for retrieval.

\subsection{Separating cross-attention from output geometry}
To separate architectural effects from output geometry, we compare Euclidean and simplex models that share cross-attention.
XAttn and ReTreever use similar shared cross-attention mechanism, but XAttn outputs Euclidean embeddings (cosine) whereas ReTreever outputs simplex embeddings (softmax, TV-based scoring).

Across datasets, XAttn retains substantial gaps comparable to other Euclidean baselines, while ReTreever achieves dramatically smaller centroid gaps and \textsc{RelGap}.
For example, on SoundDescs, XAttn has centroid $\ell_2=0.309$ and \textsc{RelGap}$=0.331$, whereas ReTreever reaches $0.012$ and $0.097$.
On Flickr30k, XAttn has centroid $\ell_2=0.527$ and \textsc{RelGap}$=0.563$, whereas ReTreever reaches $0.005$ and $0.129$.
Figure~\ref{fig:relgap_bars} shows the same trend on SoundDescs and AudioCaps: methods with Euclidean outputs (including large cross-attention models such as WAVE-7B~\cite{tang2025wave}) retain large relative gaps, while simplex outputs consistently reduce them.
This isolates output geometry as the dominant factor in gap closure, beyond any effect introduced by shared attention layers.

\subsection{Coarse-to-fine scaling: Euclidean gaps grow with dimension, simplex gaps stay small}
\label{sec:coarse}

Figure~\ref{fig:coarse} compares MRL (Euclidean coarse-to-fine) and ReTreever (simplex coarse-to-fine) across representation dimensions.
MRL exhibits increasing modality gaps as resolution/dimension grows, while ReTreever maintains small gaps across resolutions.
On AudioCaps, MRL's centroid gap grows from 0.032 at 2-d to 0.534 at 256-d ($17\times$), whereas ReTreever remains below 0.06 throughout.

\subsection{Learned simplex embeddings lie in the sparse regime}
Distance behavior on the simplex depends on the density of the probability vectors (Sec.~\ref{sec:geometry}).
We therefore estimate the Dirichlet concentration $\hat{\alpha}$ from learned embeddings (Figure~\ref{fig:theory}b).
On SoundDescs, $\hat{\alpha}$ decreases from ${\sim}18$ at $d=2$ to ${\sim}0.06$ at $d=1024$, falling below $\alpha=1$ by $d=16$.
This places high-dimensional ReTreever embeddings in the sparse regime, where baseline distances plateau rather than contract (Figure~\ref{fig:theory}a).
Despite this, the modality gap remains very small and continues to reamin very low across resolutions (Figure~\ref{fig:coarse}), and \textsc{RelGap} remains lowest for simplex methods.
This supports the interpretation that gap closure reflects improved cross-modal alignment rather than a purely geometric change in baseline distance scale.

\vspace{-0.1in}
\section{Conclusions}
\label{sec:conclusion}

We presented a geometric approach to reducing the modality gap by projecting embeddings onto the probability simplex. 
Although $S^{d-1}$ and $\Delta^{d-1}$ have the same intrinsic dimension, simplex constraints (nonnegativity and unit mass) impose a shared coordinate system that acts as cross-modal calibration.
Across five benchmarks, simplex projections reduce the centroid $\ell_2$ gap by 97--99\% while matching or improving retrieval, with \textsc{RelGap} improving as well. 
Comparisons to Euclidean cross-attention baselines (XAttn and WAVE-7B) indicate that shared cross-attention by itself does not reliably eliminate the gap, and that the simplex projection plays a central role in the reductions we observe.
Finally, coarse-to-fine analysis confirms the expected scaling behavior across resolutions and supports that gap reduction reflects alignment rather than a distance-scale artifact. 
Simplex embeddings also provide a convenient compositional and interpretable representation, and enable information-theoretic similarity measures (e.g., JS, Hellinger, KL) as alternatives to cosine distance.

\bibliography{refs}
\bibliographystyle{servicenow}

\end{document}